\documentclass[sigconf]{acmart}

\copyrightyear{2022}
\acmYear{2022}
\acmDOI{}
\acmISBN{}
\acmPrice{}
\renewcommand\footnotetextcopyrightpermission[1]{} %

\AtBeginDocument{%
  \providecommand\BibTeX{{%
    \normalfont B\kern-0.5em{\scshape i\kern-0.25em b}\kern-0.8em\TeX}}}

\usepackage{etoolbox}

\newrobustcmd{\emoji}[1]{\includegraphics[height=1.10\fontcharht\font`\A]{emoji_images/#1.png}}
\usepackage{subfig}

\usepackage[utf8]{inputenc}
\usepackage{hyphenat}
\usepackage{xspace}
\usepackage{amsmath}
\usepackage{amsfonts}
\usepackage{hyperref}
\usepackage{url}
\usepackage{booktabs}
\usepackage{multirow}
\usepackage{makecell}
\usepackage{caption}
\usepackage{minibox}
\usepackage{bbm}
\usepackage{graphicx}
\usepackage{balance}
\usepackage{mathtools}
\usepackage{color}
\usepackage{marvosym}
\usepackage{ifthen}
\usepackage{textcomp}
\usepackage{enumitem}
\usepackage{verbatim}
\usepackage{algorithm}
\usepackage{algorithmic}
\usepackage{numprint}
\usepackage{balance}

\usepackage{amsthm}
\theoremstyle{plain}

\newcommand{\chatoDisplayMode}[1]{#1}

\definecolor{MyRed}{rgb}{0.6,0.0,0.0} 
\definecolor{MyBlack}{rgb}{0.1,0.1,0.1} 
\newcommand{\inred}[1]{{\color{MyRed}\sf\textbf{\textsc{#1}}}}
\newcommand{\frameit}[2]{
  \begin{center}
  {\color{MyRed}
  \framebox[.9\columnwidth][l]{
    \begin{minipage}{.85\columnwidth}
    \inred{#1}: {\sf\color{MyBlack}#2}
    \end{minipage}
  }\\
  }
  \end{center}
}

\newcommand{\note}[2][]{\chatoDisplayMode{\def\@tmpsig{#1}\frameit{{\Pointinghand} Note}{#2\ifx \@tmpsig \@empty \else \mbox{ --\em #1}\fi}}}
\newcommand{\todo}[2][]{\chatoDisplayMode{\def\@tmpsig{#1}\frameit{{\Writinghand} To-do}{#2\ifx \@tmpsig \@empty \else \mbox{ --\em #1}\fi}}}

\newcommand{\abbrevStyle}[1]{#1}

\newcommand{\eg}{\abbrevStyle{e.g.}\xspace}
\newcommand{\cf}{\abbrevStyle{cf.}\xspace}

\newcommand{\vs}{\abbrevStyle{vs.}\xspace}
\newcommand{\etc}{\abbrevStyle{etc.}\xspace}

\newcommand{\Secref}[1]{Sec.~\ref{#1}}

\newcommand{\Tabref}[1]{Table~\ref{#1}}
\newcommand{\Figref}[1]{Fig.~\ref{#1}}

\newcommand{\xhdr}[1]{\vspace{1.7mm}\noindent{{\bf #1.}}}

\newcommand{\textcite}[1]{\citeauthor{#1} \shortcite{#1}}

\newcommand{\hide}[1]{}

\newcommand{\Emojinize}{Emojinize\xspace}

\begin{document}
\newcommand{\lars}[1]{{\color{blue} #1}}

\title{Emojinize \emoji{input-latin-letters}\emoji{right-arrow}\emoji{grinning-face}: Enriching Any Text with Emoji Translations}

\author{Lars Klein}
\affiliation{%
  \institution{EPFL}
  \city{Lausanne}
  \country{Switzerland}
}
\email{lars.klein@epfl.ch}

\author{Roland Aydin}
\affiliation{%
  \institution{TUHH}
  \city{Hamburg}
  \country{Germany}
}
\email{roland.aydin@tuhh.de}

\author{Robert West}
\affiliation{%
  \institution{EPFL}
  \city{Lausanne}
  \country{Switzerland}
}
\email{robert.west@epfl.ch}

\begin{abstract}
Emoji have become ubiquitous in written communication, on the Web and beyond.
They can emphasize or clarify emotions \emoji{grinning-face}, add details to conversations (\eg, ``I'm looking forward to the weekend \emoji{skier}''), or simply serve decorative purposes \emoji{blossom}\emoji{mushroom}\emoji{evergreen-tree}.
This casual use, however, barely scratches the surface of the expressive power of emoji.
To further unleash this power, we present \textit{\Emojinize,} a method for translating arbitrary text phrases into sequences of one or more emoji without requiring human input.
By leveraging the power of large language models, \Emojinize can choose appropriate emoji by disambiguating based on context (\eg, \emoji{cricket-game} \vs\ \emoji{bat})
and can express complex concepts compositionally by combining multiple emoji (\eg, ``\Emojinize'' is translated to \emoji{input-latin-letters}\emoji{right-arrow}\emoji{grinning-face}).
In a cloze test--based user study, we show that \Emojinize's emoji translations increase the human guessability of masked words by 55\%, whereas human-picked emoji translations
do so by only 29\%.
These results suggest that emoji provide a sufficiently rich vocabulary to accurately translate a wide variety of words.
Moreover, annotating words and phrases with \Emojinize's emoji translations opens the door to numerous downstream applications, including children learning how to read, adults learning foreign languages, and text understanding for people with learning disabilities.
\end{abstract}

\maketitle

\section{Introduction}

In today's digital age, the way we communicate is continuously evolving. While written text remains the primary mode of communication, there has been a growing inclination towards visual representations such as emoji. These colorful and intuitive symbols serve as a universal language that can span linguistic boundaries, offering a bridge between disparate cultures and generations.

The present paper explores the potential of emoji to transform the way we comprehend text, especially for specific downstream tasks enabled by a more sophisticated emoji conversion system. Consider the realm of children's books: emoji can act as helpful annotations, facilitating comprehension and even propelling early reading capabilities. Similarly, for those struggling with text comprehension due to linguistic barriers or cognitive impairments, emoji can illuminate meanings and create a semblance of universal understanding.

The task of enriching text with emoji translations is challenging. Current methods to curate emoji annotations or develop custom emoji graphics hinge on significant human intervention, often proving time-consuming and costly. Moreover, the subjective nature of emoji, combined with the ambiguity of potential interpretations, begs the question: Is an emoji language even universally comprehensible? Can humans consistently understand, and more importantly, generate meaningful sentences in this language? Can this generation by humans be outperformed by an AI-based generation method?
In this context, the present paper aims to explore several research questions:
\begin{itemize}
\item Do humans inherently understand the emoji language without explicit training?
\item Can humans effectively generate meaningful sentences using emoji, can they natively ``speak Emoji''?
\item Can an AI system rival, or even surpass, humans in enriching text with emoji translations?
\item Who excels more in the emoji translation task---humans or AI?
\end{itemize}

\xhdr{Contributions}
At the core of our study design is a cloze test, a protocol which measures the level of text comprehension by asking participants to guess a hidden word. 
We compare the guess rate between a baseline condition in which the cloze test shows only text as context for the hidden word, and conditions where an emoji language translation provides a hint for the missing word. This allows us to quantify the information gain that can be realized by annotating incomprehensible or otherwise missing text with emoji language.

We observe results both from a ``human translation'' condition, where words are translated to emoji language by human annotators, and from an ``AI translation'' condition.
To the best of our knowledge, our work is the first to introduce a scalable and fully automatic translation system for emoji language, called \textit{\Emojinize}. Unlike existing systems that merely sprinkle text with decorative emoji, \Emojinize offers true translation, considering both prior and subsequent contexts (different from next-token prediction), disambiguating synonyms based on the situation (different from a static lookup table), and harnessing the expressive power of combining multiple emoji. The system's inherent flexibility allows for it to cater to both singular words and multi-word expressions, ensuring that the translation's essence remains intact.

We consider the user study an important contribution in itself, as it provides evidence for two separate claims:

\begin{enumerate}
\item Emoji translations can strongly improve understanding.
\item Emojinize's automatic emoji translation outperforms human annotation.
\end{enumerate}

We present a fully synthetic implementation of our study protocol, emulating the replies of crowdsourced human participants with answers generated by a large language model.
This allows us to more deeply explore the capabilities of our \Emojinize translation system. We discuss a multi-shot translation mechanism that is able to integrate feedback from an oracle model into the translation mechanism and find significantly improved guess rates in the synthetic cloze test. We also study the performance of \Emojinize for multi-word expressions.

\section{Related work}

Emoji can be considered a global \cite{guardian2023emojiglobal} or ubiquitous language \cite{lu_learning_2016}.
A study on the linguistic function of emoji in social messenger communication \cite{hasyim2019linguistic} concludes that ``emoji are part of the grammatical elements of language in communicating on social media''.
In \cite{alshenqeeti_are_2016}, the authors argue that emoji are not a new language but an evolution of visual language, which can augment digital communication with new layers of meaning.
The authors of \cite{tang_emoticon_2019} conclude that people use emoji ``to avoid misunderstanding and to substitute textual expressions.''
A study on the emoji and sticker usage of thirty WeChat users confirms that in some cases emoji can replace text \cite{zhou_goodbye_2017}.
Emoji can be fun and appealing leading to their frequent usage by advertisers to increase positive affect and purchase intention \cite{das_emoji_2019}.
Data from 200 Weibo influencers shows how ``emoji rhetoric'' is used to encourage responses from followers \cite{ge_emoji_2018}.

The use of emoji allows for very intuitive communication. They may even serve as a precursor to reading, for children who don't know how to read yet \cite{wired2023childrenemoji}. Or they could be integrated as visual cues into a flashcard-based language learning system \cite{uxdesign2023emojiflashcards}
There is even a collaborative crowdsourcing project that aims to translate an entire book into emoji \cite{emojidick2023}.

The vocabulary of the `emoji language' is large, with over 3600 emoji in the latest Unicode standard \cite{emoji_statistics}.
While some emoji are well-defined, others might be ambiguous and open to interpretation \cite{miller_blissfully_2016,czestochowska_context-free_2022}.
This can lead to misunderstandings \cite{miller_understanding_2017,tigwell_oh_2016} with one reason being that emoji may be rendered and displayed differently across platforms.
Instead of different graphics on each platform, emoji could be custom-created \cite{kolla_emojify_2021}, allowing for more subtlety in expressing emotions.
To understand the meaning of emoji, it is important to consider the context in which they are used.

We describe here a mechanism by which arbitrary text passages can be translated into emoji language, and we evaluate to which extent the information contained in emoji language annotations can help with text comprehension.

A simple approach to translating a word into emoji language would be to map it to the closest emoji in the embedding space.
An embedding mechanism for emoji is described in \cite{eisner2016emoji2vec}.
However, this approach is not context-sensitive. 
In particular, services related to such static approaches such as \cite{farnum_emojipasta_2023} are not related to our work: They rely on a simple dictionary lookup to decorate text with emoji.

The authors of \cite{wijeratne_emojinet_2017} have created a large database connecting emoji with their semantic meaning. Each emoji is annotated with a set of sense labels.
Using the so-called EmojiNet database and embedding techniques to calculate the semantic similarity of emoji is done in \cite{wijeratne2017semanticemoji}.
In \cite{pohl_beyond_2017}, the authors focus on how to build an efficient emoji keyboard. For this purpose, they compute an emoji similarity measure based on emoji embeddings computed from 21 million tweets.

An LSTM neural network trained with federated learning to predict emoji is described in \cite{beaufays_federated_2019}.
This is a language model in which relevant emoji are learned with a next-token-prediction objective.
Such a model can only learn to mimic human ways of decorating text with emoji based on emoji usage in its training corpus, but it does not actually translate into emoji language.
In particular, an emoji translation based on a next-token-prediction approach can only take the preceding context into account, but not the context \textit{after} the emoji.

The state of the art in language modeling today is represented by large language models (LLMs) such as GPT-4 \cite{openai_gpt-4_2023} and Llama-2 \cite{touvron_llama_2023}, which are based on the transformer architecture \cite{vaswani_attention_2023}.
An emergent capability of these models is an intuitive grasp of the semantic meaning of emoji. This is captured as one sub-task of the BigBench benchmark \cite{srivastava2023beyond}.

LLMs can be prompted to translate between languages.
We posit that emoji language is an intuitive concept, fragments of this language can be seen sprinkled over the internet. Since LLMs are trained on large corpora crawled from the internet, they have been exposed to this language. With adequate prompting schemes \cite{wei2022chain,brown2020language}, we can tap into this latent knowledge and reveal the language.
To the best of our knowledge, there is no existing attempt at translating arbitrary text, in context, into emoji language, either with the purpose of text replacement or for annotation.

\section{Method: Translating Text to Emoji} \label{sec:translating_to_emoji}

\begin{figure}[ht]
  \centering
  \includegraphics[width=\linewidth]{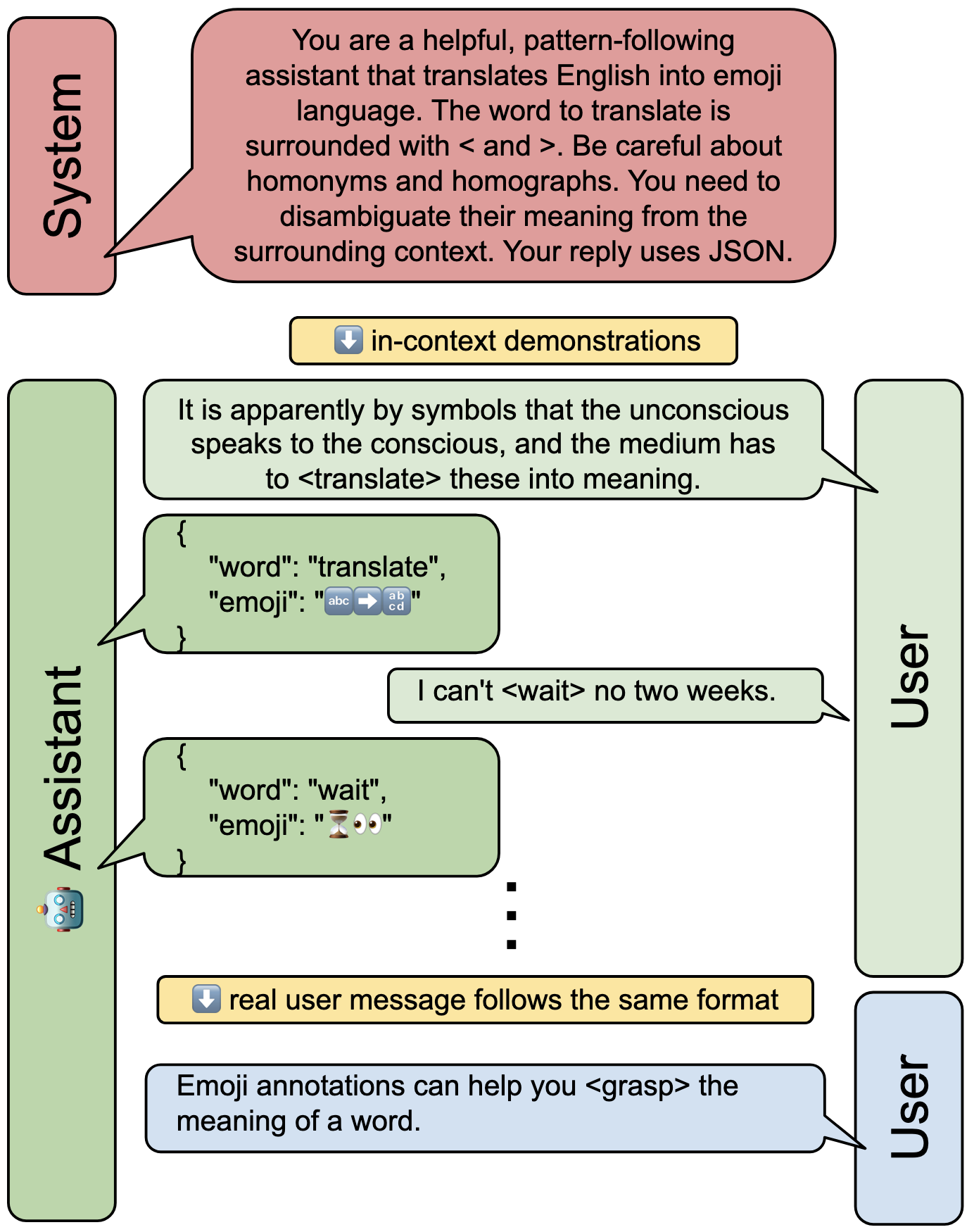}
  \caption{LLM-based translation of words to emoji using few-shot prompting with JSON-formatted assistant output.}
\label{fig:prompting_scheme}
\end{figure}

Our translation system (schematized in \Figref{fig:prompting_scheme}) relies on an understanding of emoji semantics, an emergent behavior in LLMs \cite{openai2023gpt4}. We find that the best results are achieved by GPT-4, and thus, we relied on the OpenAI API to conduct our experiments.

The task of translating a chosen text passage to emoji language is described in the system prompt. We then use few-shot learning \cite{openai2023gpt4} to guide the model towards high-quality translations. A series of in-context demonstrations each show an example text with a passage marked for translation, followed by a high-quality translation. The demonstrations are manually curated and don't intersect with the text corpus used for our experiments or the examples shown in \Tabref{tab:example_emoji_translations}. Several rounds of iteration in the prompt design led us to include a variety of scenarios in the in-context demonstrations. Relying on few-shot learning, we teach the model to choose a single emoji or to combine several emoji depending on the complexity of the text to be translated. We expose the model to text with varying writing styles, from a short snippet of slang to a long formal sentence.

Following the insights of \cite{autogpt2023} we instruct the model to output its answers in JSON format. This facilitates parsing and in particular, provides a clear criterion for rejecting incorrectly formatted replies. Rejected answers are resampled. The JSON keys are semantic and guide the model towards useful generations.

For a correct answer, the model has to fill in a given JSON template. Our template design follows the idea of chain-of-thought prompting \cite{wei2022chain}. The first part of the template repeats the text to be translated. This guides the model along a 2-step process: to first repeat the relevant text and then to translate it. 

We use a python package \cite{wurster_emoji_nodate} to filter the generated translations and to ensure that they consist only of emoji. In practice, we find that the model follows the instructions faithfully.

For an illustration of our prompting scheme, please refer to Figure \ref{fig:prompting_scheme}. It specifies the structure of our in-context demonstrations and the formatting of the JSON template used for the model replies.

Our implementation is based on a custom wrapper around the OpenAI API which adds asynchronous requests, a resource management mechanism for API key usage, and a cache. The design is heavily influenced by insights discussed in \cite{josifoski_flows_2023}. In particular, we rely on a similar caching mechanism.\footnote{Source withheld for double-blind review.}

\xhdr{Examples}
A selection of example translations is recorded in Table \ref{tab:example_emoji_translations}.
The examples have been selected to highlight the capabilities of the model:
\begin{itemize}
    \item Text is translated in context. The model can use surrounding text to disambiguate the meaning of synonyms. For a language model that proposes emoji for a next-token-prediction objective (a typical setting for mobile keyboards), only the preceding text can be taken into consideration. We can distinguish between \emoji{dove} and \emoji{building-construction} based on contextual cues from the full surrounding text.
    \item The model is aware of subtle meanings and can represent complex emotions.
    \item The translation is able to use both precise translations (there is a bat emoji, allowing a translation using only a single emoji) as well as complex compound expressions (\emoji{military-helmet} \emoji{classical-building}\emoji{counterclockwise-arrows-button} = military coup)
\end{itemize}

While the examples in Table \ref{tab:example_emoji_translations} have been manually selected, we find that they are representative of the general performance of the model. In practice, we find that there is usually no trivial improvement or superior alternative translation to the model output. This impression is confirmed by the data collected in a comprehensive user study and discussed in Section \ref{sec:eval_results}. Please refer to Section \ref{sec:exploration} for a further discussion of the capabilities of our translation mechanism.

Before collecting data for the user study presented in Section \ref{sec:experimental_results} we performed a small-scale ablation test. We found that lower temperatures worked well for this task and conducted all our experiments with a temperature of 0.

\begin{table}[ht]
  \caption{Examples of Emojinize's emoji translations.}
  \label{tab:example_emoji_translations}
  \begin{tabular}{cl}
    \toprule
    Input text&Emoji translation\\
    \midrule
    The player was ready, the <bat>\\ whooshed as it swung by. & \emoji{baseball}\emoji{cricket-game}\\[0.5em]
    The scout was ready, the <bat>\\shrieked as it flew by & \emoji{bat}\\
    \midrule
    The wall is <blue> & \emoji{blue-circle}\\[0.5em]
    I'm feeling <blue> today & \emoji{disappointed-face} \emoji{blue-heart}\\
    \midrule
    In the distance we saw a giant\\ <crane> fly over the swamp. & \emoji{dove} \\[0.5em]
In the distance we saw a giant\\ <crane> tower over the dump. & \emoji{building-construction} \\
\midrule
The military <coup> had greatly\\ destabilized the trust of\\ foreign investors. &  \emoji{military-helmet} \emoji{classical-building}\emoji{counterclockwise-arrows-button}\\
  \bottomrule
\end{tabular}
\end{table}

\section{Evaluation protocol: cloze tests}
\label{sec:experimental_results}

\begin{figure*}[ht]
    \centering
    \includegraphics[width=\textwidth]{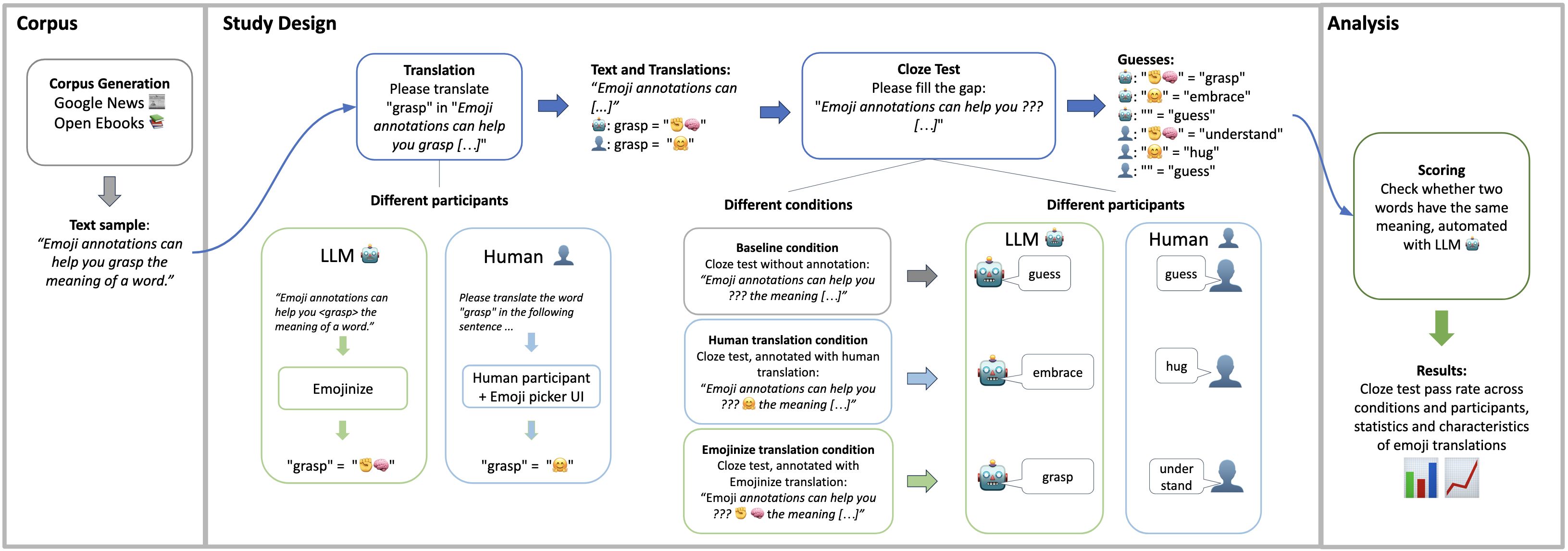}
    \caption{Evaluation protocol. Text from a custom corpus is translated to emoji language. Translations are used in a cloze test. Different translation mechanisms lead to different test conditions. For the analysis we rely on LLM-based scoring.}
    \label{fig:study_design}
\end{figure*}

Our study is designed to measure the information contained in emoji translations.
We start with the question: Can emoji annotations help with understanding text, and if so, can we quantify that effect?

A cloze test is a test of comprehension in which some words in a text passage are removed and must be guessed by the reader. The percentage of correctly guessed words (accuracy) is a proxy measure for text comprehension.

We create a text corpus of small text snippets (see Section \ref{sec:data}). Our baseline is a cloze test based on this corpus, with a random word selected and hidden in each phrase.
The ``human annotation'' condition displays an emoji translation created by a human as a hint for the hidden word.
In the ``\Emojinize annotation'' condition, we use a translation created by \Emojinize instead.

To decide whether the answers in the cloze test match the hidden word, we rely on an LLM and ask whether two given words have the same meaning. This allows us to detect synonyms and ignore small spelling mistakes. We explicitly instruct the model to ignore typos.

We evaluate 1000 text samples in the baseline, human annotation, and \Emojinize annotation conditions, leading to a total of 3000 data points.
To first collect human emoji translations and then conduct the cloze test, we organize two crowdsourced user studies. For details on these studies, please refer to Section \ref{sec:human_eval}. 
An automatic evaluation of our study design, which mimics the crowdsourced setup but relies on LLM ``participants'' is discussed in Section \ref{sec:automatic_eval}.

Our study design is explained visually in Figure \ref{fig:study_design}.

\subsection{Human evaluation} \label{sec:human_eval}

We first conduct a crowdsourced study to gather human translations. 
One translation is collected for each text sample.

To ensure high-quality data, we must control the input mechanism by which participants select an emoji. If we simply display a text input area and ask the participant to enter an emoji translation, they would rely on the system's native emoji picker. This can be different across devices. Many systems also include convenience features such as a highlight reel of frequently or recently used emoji or a search mechanism (typically based on a simple keyword search). This would introduce systematic bias. We instead want to enable the participant to use the full expressive power of all emoji in the Unicode standard. 

It would be possible to use a variety of open-source emoji picker widgets. This introduces some degree of control and would allow us to switch off features such as keyword search. However, to the best of our knowledge, all high-quality emoji pickers still implement some form of categorization and keep emoji in tabs such as ``sports'', ``plants'' etc. They are also designed to be space efficient and do not leverage the entire screen area of a desktop PC or laptop.

\begin{figure*}[h]
  \centering
  \includegraphics[width=0.8\linewidth]{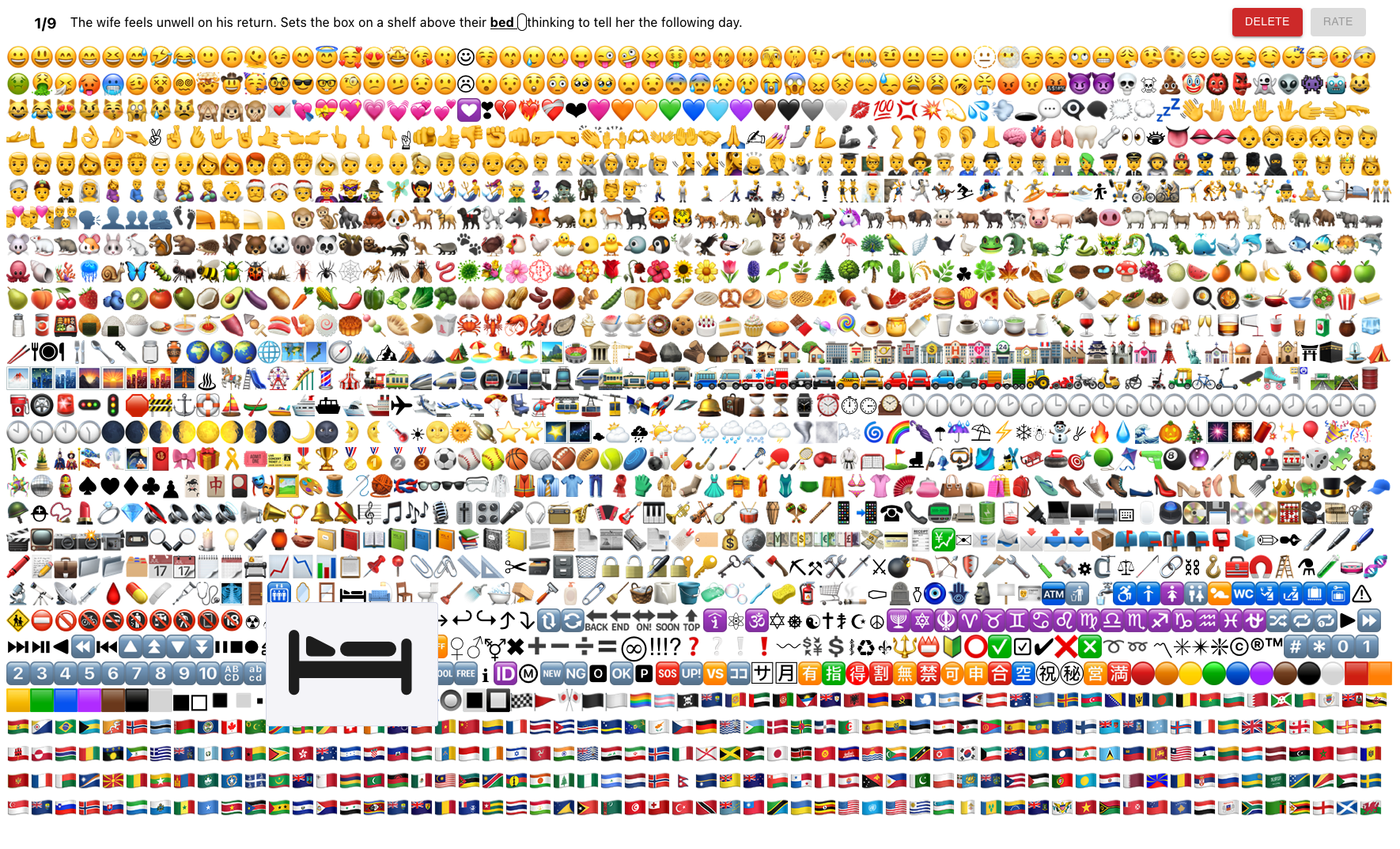}
  \caption{A custom emoji picker for allowing humans to translate target words (here: ``bed'') to emoji (\Secref{sec:human_eval}).}
\label{fig:emoji_picker}
\end{figure*}

\begin{figure}[h]
  \centering
  \includegraphics[width=\linewidth]{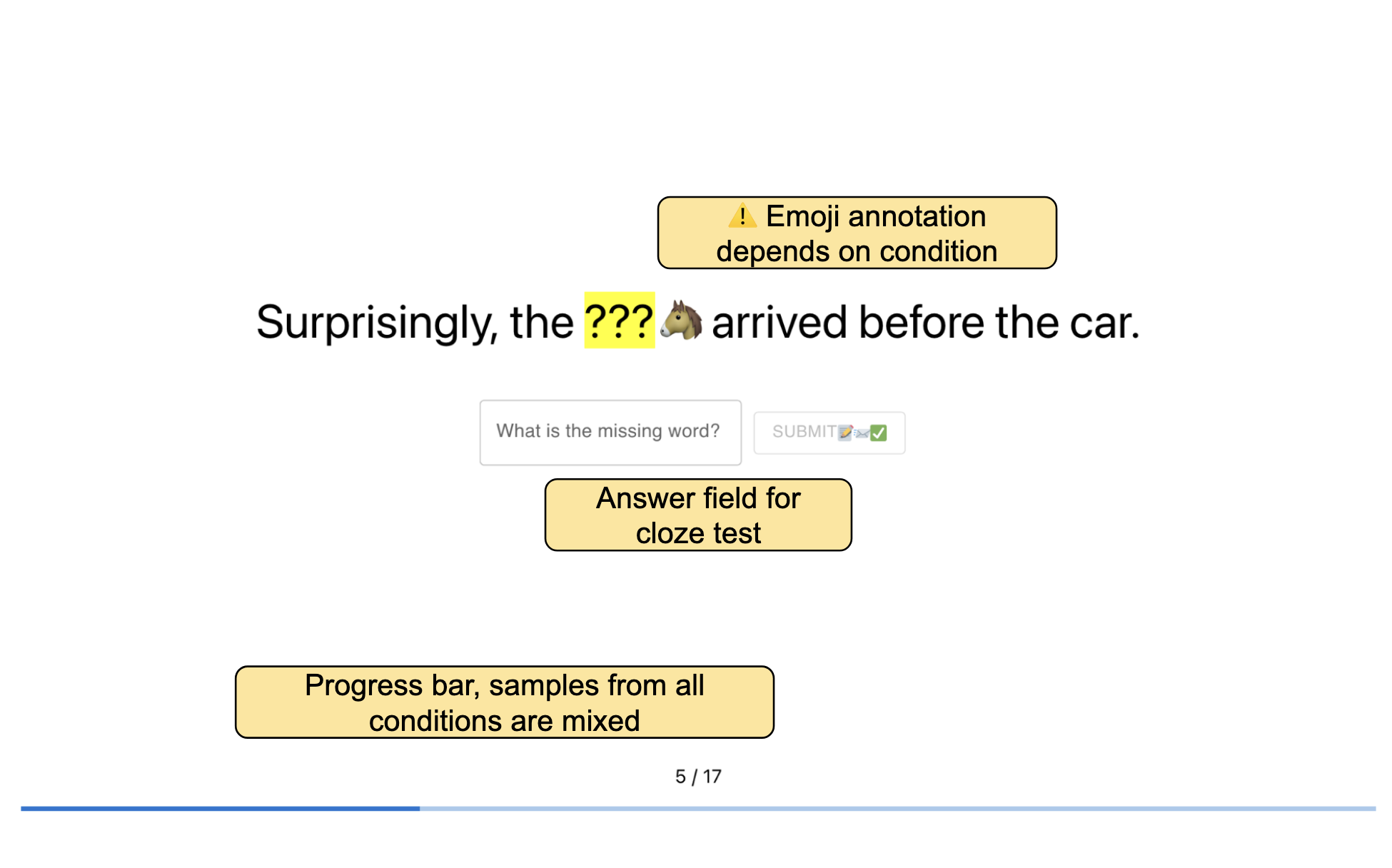}
  \caption{UI of cloze test for human evaluation.}
\label{fig:e2t}
\end{figure}

We hence decided to build a custom emoji picker widget. It displays all emoji in a grid, which is automatically adjusted to fill the entire user screen. When hovering over an emoji, it is enlarged. Please refer to \Figref{fig:emoji_picker} for a screenshot of our emoji picker implementation. \Figref{fig:e2t} shows the cloze test UI for human guessing.

\subsection{Automatic evaluation}\label{sec:automatic_eval}

Conducting the user study described in Section \ref{sec:human_eval} is costly and time-intensive.
We therefore propose to also emulate the answers of crowdsourced human participants with replies from an LLM.
This allows us to explore the capabilities of \Emojinize and to evaluate a variety of extensions more quickly and cheaply.

In this setup, for an automatic evaluation of emoji translations, we implement a prompting scheme that sets up an LLM to answer the questions in our cloze test. Similar to \Secref{sec:translating_to_emoji}, we use few-shot learning and in-context demonstrations. In this case, we don't use JSON formatting for the generated answers and instead instruct the model to simply reply with its guess for the hidden word.

\subsection{Data} \label{sec:data}

For our study, we created two new corpora, designed to cover a broad range of topics, writing styles, and diverse levels of complexity.
Both corpora only contain text that was written after the knowledge cutoff of GPT3.5-Turbo and GPT4. This ensures that our automatic evaluation (\Secref{sec:automatic_eval}) is not contaminated by information seen during pretraining.

The first corpus is based on news articles featured on the Google News portal \cite{noauthor_google_2023}. Starting from diverse queries such as "Environmental conservation initiatives", "Positive news stories", "Inspiring stories of human achievement", \etc, we collect a selection of news articles. We then download the HTML of each article and extract all text contained in 
paragraph tags (<p></p>).
The second corpus is based on 5 open and freely available ebooks \cite{noauthor_paycheck_nodate, noauthor_ghostly_nodate, noauthor_legend_nodate, noauthor_making_nodate, noauthor_oldest_nodate}.

From each of the corpora, we sample random text passages and apply a rigorous series of cleaning steps. As a preliminary filter, the text is lemmatized \cite{honnibal2020spacy} and compared against a list of profanities.
We, however, find that this is not sufficient: sentences scraped from news articles may contain spam or advertisement, such as the request to subscribe to a newsletter, be in a language other than English, or have a variety of formatting issues.
Moreover, sentences taken from a book may contain violent or sexual content, often described in an implicit and indirect way. Due to their diverse vocabulary and creative use of imagery, these descriptions can avoid detection in the lemmatization-based check against profanities.

We thus prompt an LLM to detect if any such problems exist in a candidate text. Only text passages that pass all quality criteria are selected.
We do not manually include or exclude any text, we only tune the LLM-based filter until all random samples that pass the checks are free of problematic content.

Finally, we randomly sample a noun, verb, adjective, or adverb from each of the selected sentences. Punctuation, proper nouns, stopwords, \etc, are filtered out. The tokenization and POS-tagging are performed with spaCy \cite{honnibal2020spacy}.

The final corpus comprises 1000 samples, 500 from articles featured on Google News and 500 from open ebooks. The samples are shuffled in random order.

\section{Evaluation results}\label{sec:eval_results}

\begin{figure}
     \centering
     \subfloat[][Human guesses hidden word]{\includegraphics[scale=0.4]{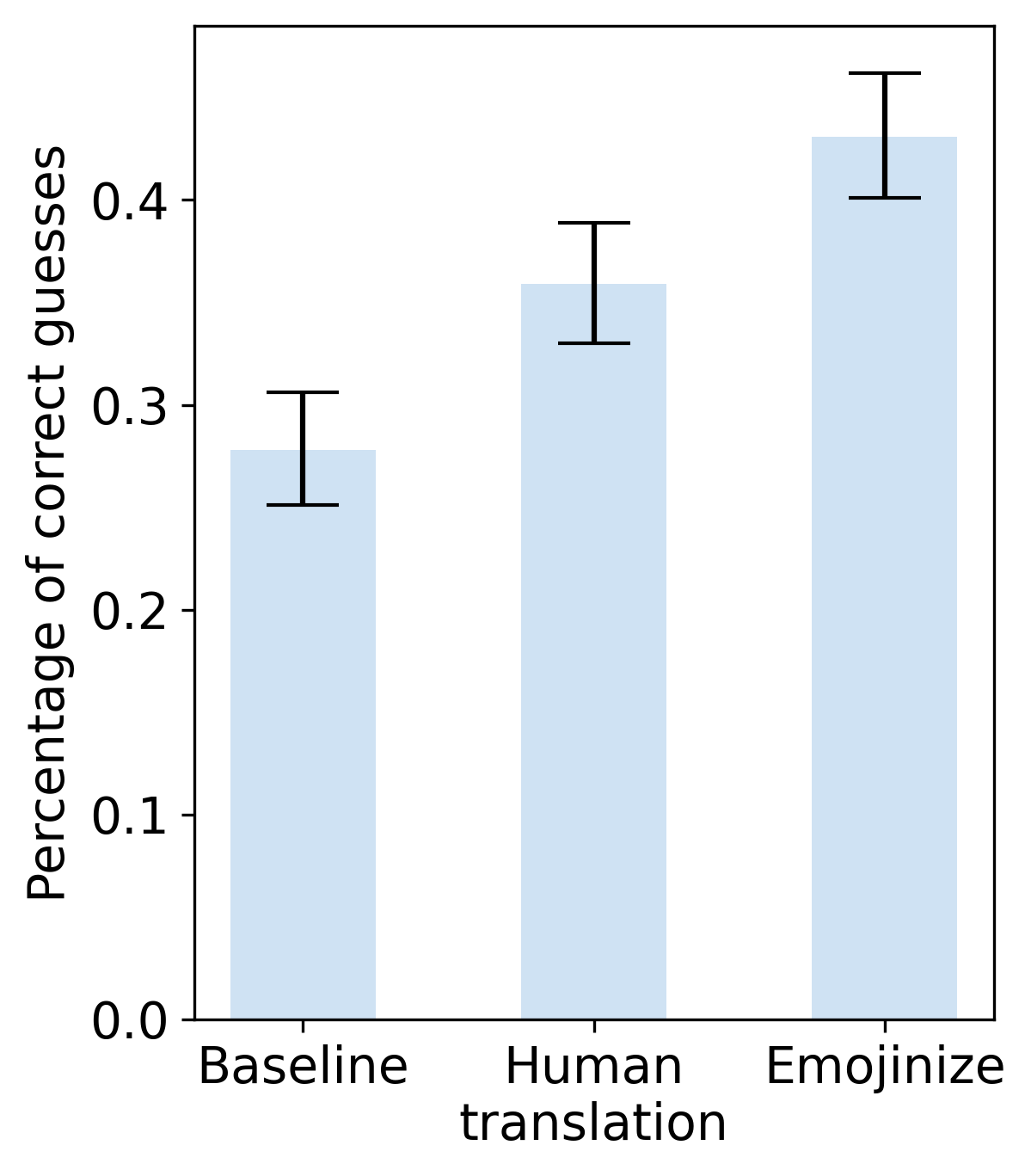}\label{<figure1>}}
     \subfloat[][LLM guesses hidden word]{\includegraphics[scale=0.4]{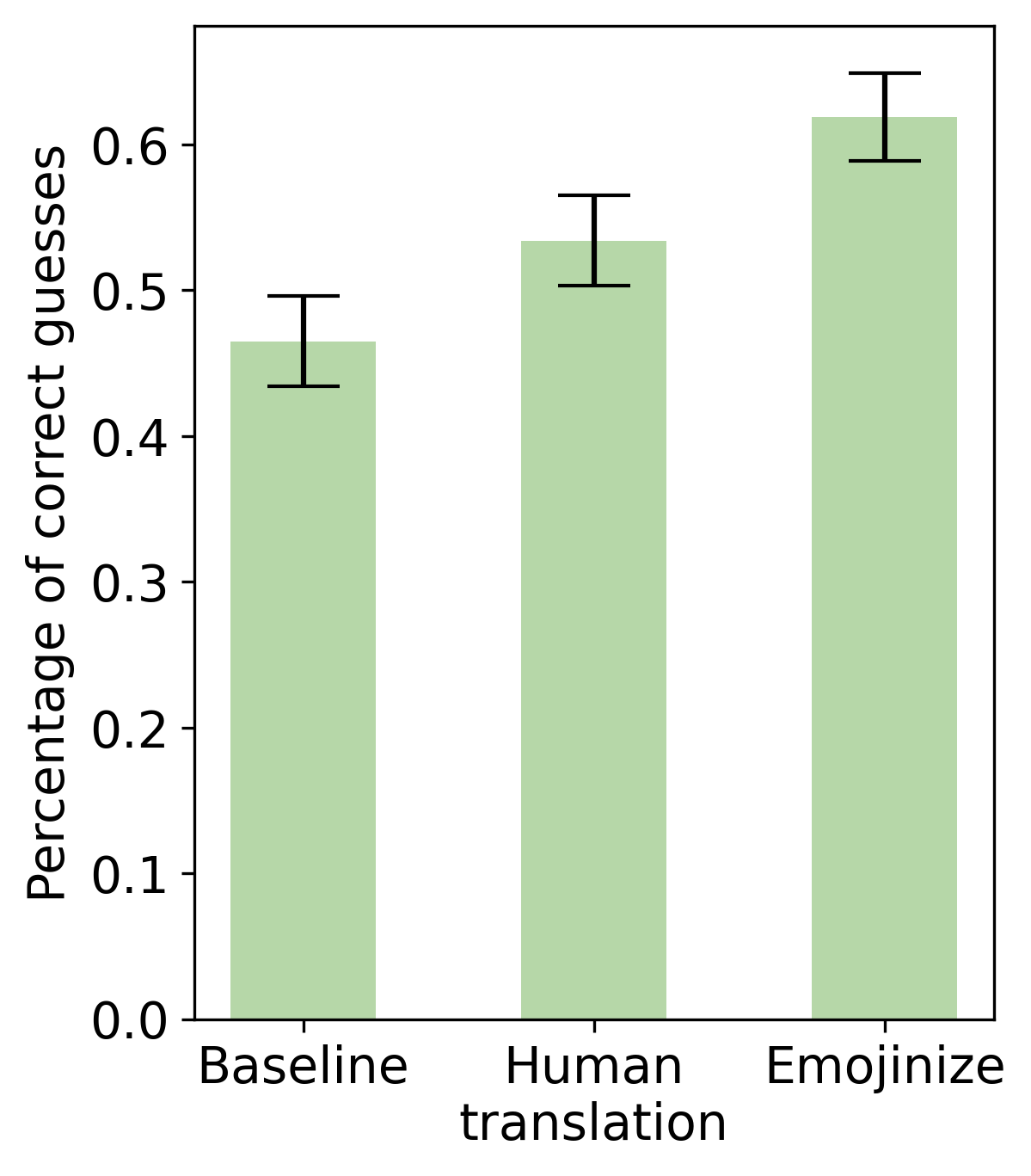}\label{<figure2>}}
     \caption{Accuracy of guessing the hidden word: (a)~human participants (\Secref{sec:Human results}), (b)~LLM participants (\Secref{sec:Automatic results}).}
     \label{fig:human_llm_eval}
\end{figure}

For the human evaluation, we conducted a crowdsourced user study following the protocol described in \Figref{fig:study_design}.
In total, we collected data from 508 participants.
As we shall see in what follows (\Secref{sec:Human results}), the data yields clear and statistically significant results for our research questions:
Without any prior training, human participants understand the emoji translations and use them to vastly improve their guess rate in the cloze test.
The automatic evaluation (\Secref{sec:Automatic results}), where an LLM replaces the human guesser, yields the same conclusion.

\begin{table}
  \caption{Accuracy of guessing the hidden word (plot in \Figref{fig:human_llm_eval}).}
  \label{tab:userstudy}
  \begin{tabular}{cll}
    \toprule
    Participant&Condition&Accuracy\\
    \midrule
    Human&Baseline & 0.278, [0.251, 0.306]\\
    Human&Human translation & 0.359, [0.330, 0.389]\\
    Human&\Emojinize & \textbf{0.431, [0.401, 0.462]}\\
    \midrule
    LLM & Baseline & 0.465, [0.434, 0.496]\\
    LLM & Human translation & 0.534, [0.503, 0.565]\\
    LLM & \Emojinize& \textbf{0.619, [0.589, 0.649]}\\
  \bottomrule
\end{tabular}
\end{table}

\subsection{Human evaluation}
\label{sec:Human results}

In the human evaluation, we compare translation generated by humans to those generated by \Emojinize, and to the no-emoji baseline.

The results displayed in \Figref{fig:human_llm_eval} and \Tabref{tab:userstudy} show that,
whereas in the no-emoji baseline condition, humans guess the hidden word correctly 27.8\% of the time,
the emoji translations produced by human annotators are helpful by increasing the accuracy to 35.9\%---a relative improvement of 29\%. 
\Emojinize translations are better still, with a relative improvement of 55\% (nearly twice as large as 29\%), to an accuracy of 43.1\%.

The non-overlapping 95\% confidence intervals of \Figref{fig:human_llm_eval} imply  that the differences in guessability of baseline, human, and \Emojinize translations are statistically significant at $p<0.05$.

\subsection{Automatic evaluation}
\label{sec:Automatic results}

We use the same cloze test as in the human evaluation but replace crowdsourced participants with an LLM.
The baseline performance of the LLM is 46.5\%.
Annotating a hidden word with human-generated emoji translations increases the accuracy to 53.4\%.
With \Emojinize, the accuracy increases to 61.9\%.
Again, the 95\% confidence intervals do not overlap, implying that the results are statistically significant with $p<0.05$.

We analyze a correlation between the correct guesses of the LLM and the human participants.
The correlation coefficient is 0.324.
With access to a highly reliable oracle that predicts the probability of a human correctly guessing the hidden word, 
we could further improve the translation quality. This approach is discussed in Section \ref{sec:multi_shot}.

\subsection{Properties of human \vs\ Emojinize translations}

The average length of a human emoji translation is 1.487 emoji, with a 95\% confidence interval of [1.425, 1.553].
In total, human participants used 660 distinct emoji. We calculate the relative frequencies of each emoji in the generated translations, the entropy of this categorical distribution is 6.164.
For comparison, the average length of Emojinize's translations is 1.777 emoji (95\% CI [1.732, 1.827]), significantly longer than human translations (see above).
At the same time, we find that \Emojinize leads to a lower entropy in the distribution of emoji.
The entropy of the distribution of emoji is 5.582, with 471 unique emoji being used at least once.

\section{Exploring additional capabilities} \label{sec:exploration}

 \subsection{Multi-shot Emojinize} \label{sec:multi_shot}

Some concepts can only be translated by combining the meaning of multiple emoji.
As the expressions become more complex, this places a burden on the reader to correctly interpret the emoji translation produced by the model.
There is a tradeoff to be made between the range of language that can be represented with emoji language and the risk of being misunderstood.

Next to the uncertainty of interpreting more sophisticated expressions in emoji language, there are other reasons for potential miscommunication. Differences in cultural background can lead readers to associate a different meaning with an emoji \cite{lu_learning_2016}. Varieties in emoji representation and rendering also contribute to different emoji language dialects across platforms such as Android or iOS \cite{tigwell_oh_2016}.

We propose to integrate a utility function that can provide feedback for the emoji translation process. 

The translation process based on LLMs is a stochastic process. When setting the temperature to 0, it is possible to greedily decode the most likely token at every step of the output generation. But we can also increase the temperature and explore a variety of possible translations.
This allows us to generate a list of potential translation candidates and present each of them to the oracle.

To study the potential gains in accuracy, we look at a synthetic experiment, where we integrate a backtranslation step into the emoji translation. We use an LLM to guess the original text for every translation candidate. We sample multiple independent guesses and calculate the probability of guessing correctly. This serves as a utility function. We rely on the same mechanism to evaluate the translation but generate a new, independent sample. This is similar to a crowdsourcing study where the translation of a text passage is generated in an iterative fashion, taking into consideration the feedback from a set A of crowdsourcing workers. After deciding on a final translation, the translated text is evaluated by a separate, disjoint set B of crowdsourcing workers.

\begin{figure}[h]
  \centering
  \includegraphics[width=\linewidth]{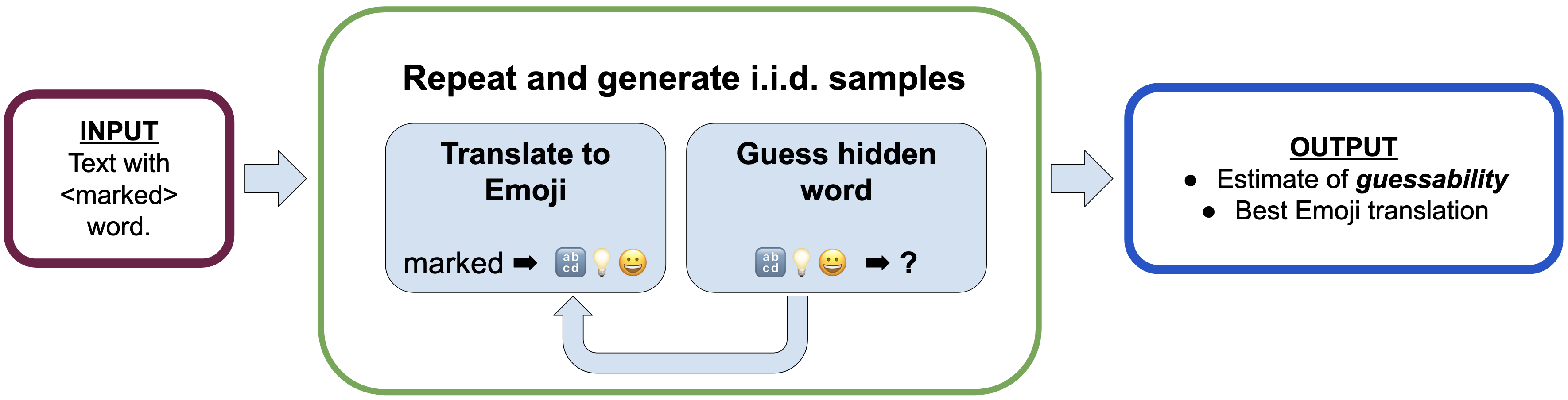}
  \caption{Simulating a human reader to choose the best possible emoji translation and estimate its guessability.}
\label{fig:multi-shot}
\end{figure}

The concept of this multi-shot algorithm is explained in Figure~\ref{fig:multi-shot}.

We have conducted a synthetic experiment and found strong, statistically significant improvements, pushing the guess rate up to 73.4\%, from 61.9\%. Results are recorded in Figure \ref{fig:multi_shot_and_word}.

In our scenario, both the answers for translation improvement and final evaluation are sampled independently from the same distribution. This means the feedback during the translation phase is optimal since any change in the distribution of feedback between the translation and evaluation phases could result in a misalignment of objectives and might even lead to a decrease in translation quality.

Ideally, we would need access to an oracle that accurately models the probability of a human reader understanding the emoji translation. The utility function we have created here relies on the probability of an LLM correctly guessing the hidden word. It is possible to use this as a proxy measure for the probability of a human reader correctly guessing the hidden word.

It is questionable whether an LLM is an accurate model of human understanding. Indeed, even the baseline probability of an LLM passing the fill-the-gap test without any annotation is larger than the probability of a human reader correctly guessing the hidden word with a high-quality LLM translation. 

Nevertheless, we have included multi-shot translations in our user study. The fill-the-gap guess rate of human participants with multi-shot emoji translations is 42.6\% [0.396, 0.456]. The CI doesn't overlap with the baseline or human translations. However, compared to single-shot translations, the guess rate is slightly worse, and the difference is not statistically significant.

\subsection{Emojinizing multiple words in one go}

The implementation of \Emojinize presented in Section \ref{sec:translating_to_emoji} translates one marked text passage per input document.
If the user wishes to translate multiple passages, \Emojinize can be run multiple times.

It is possible to extend this 1:1 relation between the input document and output translation to support an arbitrary number of translations for each input text. 
This increases sample efficiency and leads to a speedup factor of N if N translations are produced in one batch. Particularly when processing large corpora such as books, the cost of inference is an important consideration.

With an adjustment to the prompt formatting, \Emojinize can translate an arbitrary number of marked passages for each input.
We use the same prompting setup as demonstrated in Figure \ref{fig:prompting_scheme}; in particular, we reuse the same demonstration texts. But adjust the marking of input texts and the JSON format to include multiple translated words.

We repeat the automatic evaluation of our user study and find that the translation performance is slightly improved. When translating 3 random words per phrase, the guess rate is 0.635 [0.618, 0.652], compared to 0.619 [0.589, 0.649] for a single translation.

The small and non-statistically significant improvement could be caused by an increased number of emoji language examples in the demonstrations. We keep the same number of demonstrations and the same texts but now include several marked passages and reference translations for each text.

\subsection{Multi-word expressions}
\label{sec:mwe}

A Multi-Word Expression (MWE) is a phrase containing more than one word. For instance, commonly known MWEs include phrases such as "social security" or idiomatic expressions like "spill the beans". 

We can identify MWEs by prompting a large language model. In the context of \Emojinize, we can use the versatility of such a prompting scheme to go beyond only identifying MWEs and instruct the model to find all possible translation units. This can be any complex expression, idiom, or text phrase as long as it consists of one consecutive string of words.

\begin{figure}
     \centering
     \subfloat[][Multi-shot Emojinize]{\includegraphics[scale=0.4]{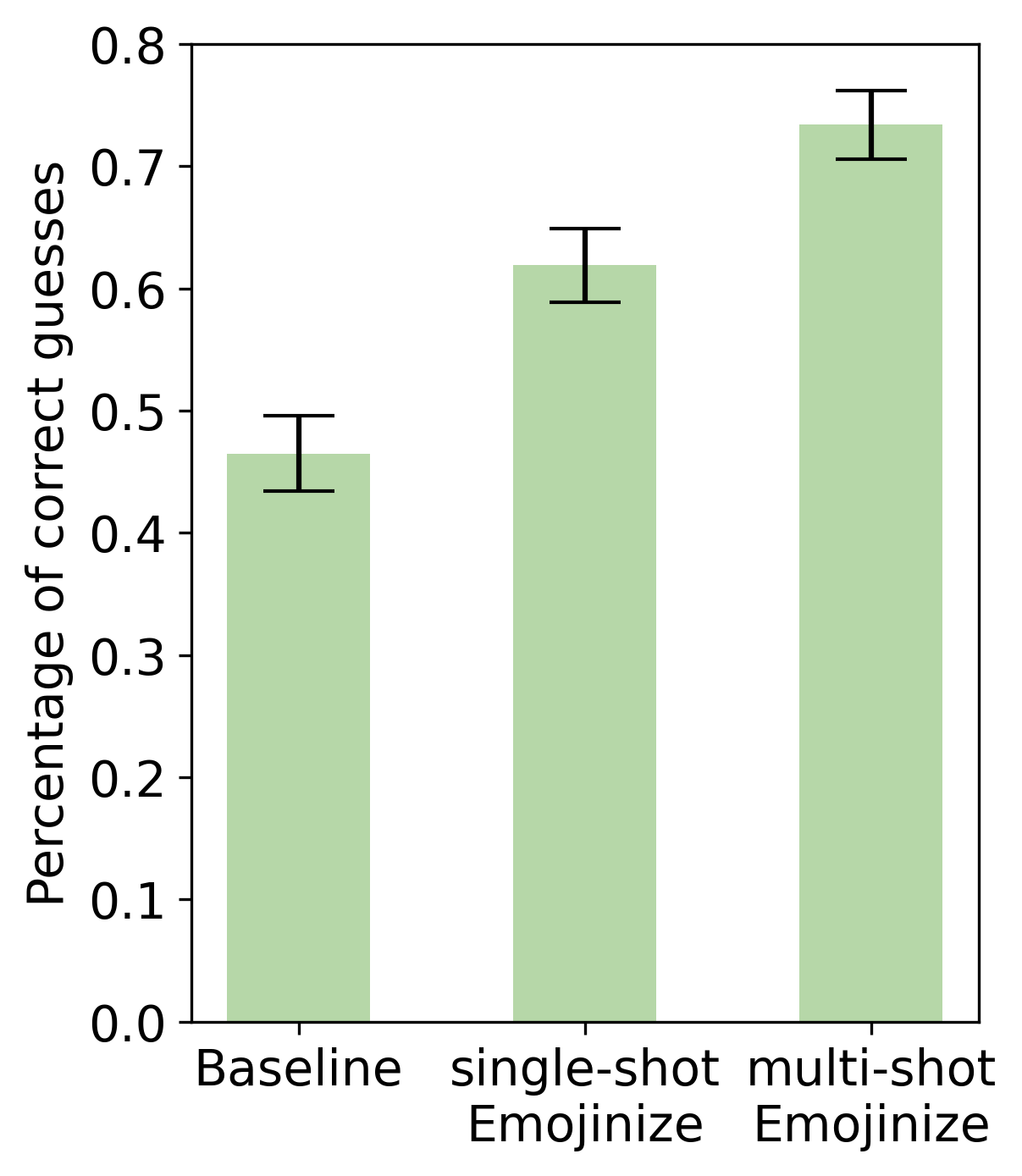}}
     \subfloat[][Multi-word expressions]{\includegraphics[scale=0.4]{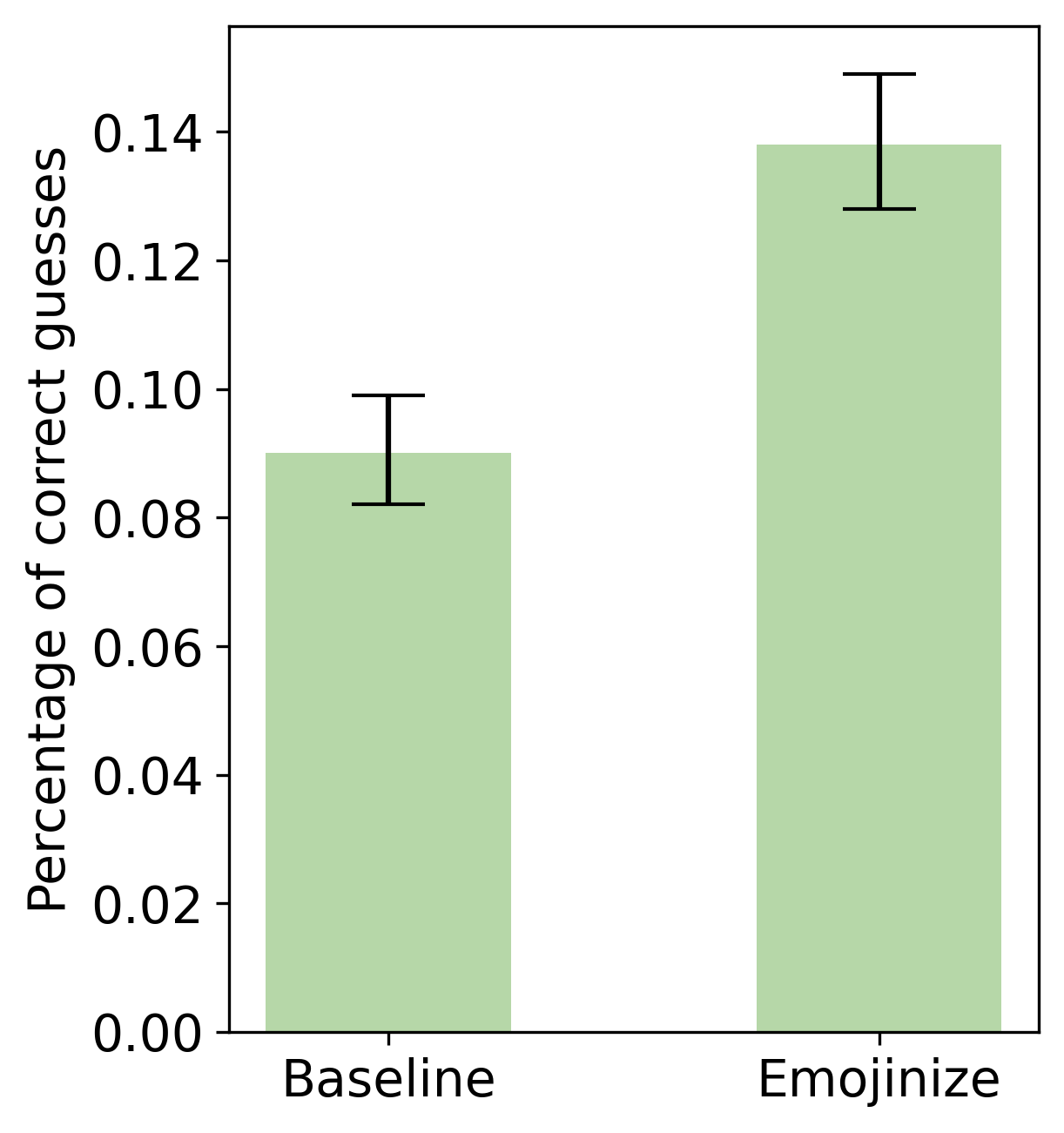}}
     \caption{Performance of (a) multi-shot Emojinize (\Secref{sec:multi_shot}), (b) Emojinize for multi-word expressions (\Secref{sec:mwe}). Since the cloze tests performed here are of different difficulty, the y-scales differ.}
     \label{fig:multi_shot_and_word}
\end{figure}

We adjust the demonstrations of \Emojinize and include examples of multiple consecutive words being translated. Then we repeat our study. Results are recorded in Figure \ref{fig:multi_shot_and_word}, the baseline pass rate drops to 0.09 [0.082, 0.099], with annotations produced by \Emojinize this is increased to 0.138 [0.127, 0.148].

It is important to note that correctly guessing multiple consecutive words is exponentially harder than guessing just a single hidden word. While the rate of correct guesses with \Emojinize annotations is at only 0.138, this still represents a relative increase of 53\%.

\subsection{Languages other than English}

The participants of our user study are filtered on English proficiency. 
The corpus study contains only English phrases. 
We have experimented with other languages and find that \Emojinize performs very well there, too.
The examples recorded in \Tabref{tab:other_language_table} were created without any adjustment to \Emojinize (except for changing ``English'' to ``German'' or ``French'' in the system prompt, \cf\ \Figref{fig:prompting_scheme}), showing that it is possible to translate German or French input text into emoji language even when using English in-context demonstrations.
At the same time, the system retains its capabilities to:
\begin{itemize}
    \item disambiguate the meaning of a word from context. In German ``blue'' is a slang expression for ``drunk''. \Emojinize picks up on the context clues that here the German slang, as opposed to the English slang is being used.
    \item combine multiple emoji into a descriptive compound expression. The word ``clearing'' can be very accurately represented by a sunny area between trees.
\end{itemize}

\begin{table}[h]
  \caption{Emojinize applied to German/French inputs.}
  \label{tab:other_language_table}
  \begin{tabular}{cl}
    \toprule
    Input Text&Emoji translation\\
    \midrule
    Das Haus ist <blau>. \\ \textit{The house is <blue>.} & \emoji{large-blue-circle}\\[0.5em]
    Der Autofahrer ist <blau>. \\ \textit{The driver is <drunk>.}& \emoji{beer-mug}\emoji{wine-glass}\emoji{woozy-face}\\
    \midrule
    Wir spielen <Federball>.\\ \textit{We're playing <badminton>.}& \emoji{badminton}\\[0.5em]
    \midrule
    Der Hirsch steht auf der <Lichtung>.\\ \textit{The stag is standing in a <clearing>.}
 & \emoji{deciduous-tree}\emoji{deciduous-tree}\emoji{sun-with-face}\emoji{deciduous-tree}\emoji{deciduous-tree} \\[0.5em]
    \midrule
    Je ne peux pas <attendre> deux semaines.\\ \textit{I can't <wait> two weeks.}
 & \emoji{hourglass}\emoji{eyes} \\[0.5em]
  \bottomrule
\end{tabular}
\end{table}

\section{Discussion}

Emoji can add an emotional layer to textual communication \emoji{smiling-face-with-hearts} and are useful to add a little bit of color \emoji{blossom}.
Sometimes, they're even the most salient part of communication, ``coming by \emoji{car}'', ``\emoji{heart} you'', ``let's have \emoji{pizza}''.

When combined creatively, the over 3600 emoji in the Unicode standard can express a wide range of ideas and concepts.
Emoji language is a fascinating phenomenon -- a completely decentralized, emergent language.
It is based on the intuitive visual semantics of emoji: You don't need a dictionary to understand emoji language.
Our comprehensive user study shows that emoji annotations convey information that can help a human reader better understand text, with a 55\% increase of correct guesses in a cloze test.
Most importantly, apart from a short introduction text explaining our study setup, the participants didn't receive formal training.
Nevertheless, they could immediately leverage the information from emoji annotations to improve their text comprehension.
Everyone is a ``native speaker'' of emoji language.

Fragments of emoji language are sprinkled across all manners of human communication.
However, emoji are chosen opportunistically based on the users' mobile keyboard suggestions, the impromptu conventions of a group chat, or the whims of a social media post.

While humans can intuitively understand emoji language, creating new translations from text to emoji language is challenging and time-consuming.
It is now well known that Large Language Models (LLMs) have displayed competence in translating between distinct languages such as English, French, etc. 
This translation ability stems largely from the model's exposure to diverse linguistic corpora. 
Despite the emoji language's fragmented and decentralized occurrence on the web, recent large language models show an intuitive grasp of emoji semantics.
We leverage this ability to create a new algorithm, \Emojinize, which translates English text into emoji language.
Our study shows that translations generated by \Emojinize are significantly more accurate than translations created by human annotators.

\Emojinize eloquently combines multiple emoji into complex compound expressions.
Nevertheless, in some areas, the vocabulary of emoji language lacks the precision of natural language.
We can represent ``dog'' but not ``spaniel'', ``bird'' but not ``heron''. 
Even in these examples, emoji annotations can still be useful. 
To a reader who doesn't know the word ``spaniel'', the emoji annotation ``\emoji{dog}'' will still provide relevant information, i.e., that the word refers to a type of dog.

The generality of \Emojinize is demonstrated in its Multi-Word Expressions (MWEs) handling. While the performance on these MWEs does degrade, there's an unmistakable surge in performance with \Emojinize, increasing user guessability of masked words. 
This speaks to the method's capability to work gracefully under complex linguistic scenarios, demonstrating a relative increase of 53\% in correct guesses even when handling more challenging multi-word translations.

\Emojinize realizes the largely untapped potential of emoji language.
We already see impressive capabilities, but as of today, the language has to be considered still in its infancy. Its evolutionary trajectory can be likened to a feedback loop: as emoji become more prevalent, awareness of their meaning and potential uses increases. This rise in awareness fuels an ongoing consensus process wherein standardized emoji usages and meanings get established and updated. With Emojinize’s ability to significantly outperform human translators both in speed and translation accuracy, access to high-quality emoji translations could become a commodity. This could catalyze the aforementioned feedback process and lead to an even faster adoption of emoji language.
Moreover, this feedback mechanism is not limited to strictly \textit{human}-emoji interaction: The more emoji are used in diverse contexts, the richer the training data becomes, and the better the LLM becomes at discerning and generating appropriate emoji translations. This creates a positive reinforcement loop, fostering natural selection in translations wherein efficient translations are recycled and reused more frequently.

Future research directions could pivot toward understanding how humans perceive and interpret emoji. If this human understanding can be mapped effectively to an LLM, which seems to be the current trend, it opens the door to shifting the entire language creation process for an emoji language to synthetic, LLM-generated data. We have already shown the potential of a multi-shot translation mechanism. An oracle model that provides feedback aligned with the human interpretation of emoji language can vastly improve translation accuracy and will help to prevent misunderstandings. 

Reliable, high-quality emoji translations can be useful in a variety of applications.
Notably, the visual and immediately intuitive semantics of this language make it possible to communicate information even to illiterate users.
In the landscape of literacy education, the challenge remains how to bridge the gap between illiteracy, whether concerning young children or adults, and reading proficiency, especially for learners who may find traditional alphabetic scripts intimidating or inaccessible. Emoji, with their pictorial representation and universality, offer a promising intermediate step in this learning curve. By nature, emoji are intuitive and often represent tangible emotions or objects, making them easier to comprehend than abstract letters or words. Thus, associating words with corresponding emoji can serve as a mnemonic aid, facilitating the memorization and recognition of words. Moreover, emoji add an element of fun and relatability to the learning process, which serves as a motivating factor. Instead of grappling with abstract text, learners can first relate to the familiar visuals of emoji, gradually transitioning to the associated text as their confidence and skills grow.

\bibliographystyle{ACM-Reference-Format}
\bibliography{bibliography}

\appendix

\end{document}